\documentclass[10pt, a4paper]{article}
\usepackage{lrec}
\usepackage{multibib}
\newcites{languageresource}{Language Resources}
\usepackage{graphicx}
\usepackage{tabularx}
\usepackage{soul}
\usepackage{epstopdf}
\usepackage[utf8]{inputenc}
\usepackage{hyperref}
\usepackage{xstring}
\usepackage{arabtex}
\usepackage{color}
\usepackage{booktabs}
\novocalize

\newcommand{\Ni}{({\em i})~}
\newcommand{\Nii}{({\em ii})~}
\newcommand{\Niii}{({\em iii})~}

\title{AraBERT: Transformer-based Model for\\Arabic Language Understanding}

\name{Wissam Antoun*\thanks{*Equal Contribution}, Fady Baly*, Hazem Hajj}

\address{American University of Beirut \\
         \{wfa07, fgb06, hh63\}@aub.edu.lb\\}

\abstract{
The Arabic language is a morphologically rich language with relatively few resources 
and a less explored syntax compared to English. Given these limitations, Arabic Natural Language Processing (NLP) tasks like Sentiment Analysis (SA), Named Entity Recognition (NER), and Question Answering (QA), have proven to be very challenging to tackle. Recently, with the surge of transformers based models, language-specific BERT based models have proven to be very efficient at language understanding, provided they are pre-trained on a very large corpus. Such models were able to set new standards and achieve state-of-the-art results for most NLP tasks. In this paper, we pre-trained BERT specifically for the Arabic language in the pursuit of achieving the same success that BERT did for the English language. The performance of AraBERT is compared to multilingual BERT from Google and other state-of-the-art approaches. The results showed that the newly developed AraBERT achieved state-of-the-art performance on most tested Arabic NLP tasks. The pretrained araBERT models are publicly available on \href{ https://github.com/aub-mind/araBERT}{github.com/aub-mind/araBERT} hoping to encourage research and applications for Arabic NLP.
\\\newline
}
\begin{document}
\maketitleabstract
\section{Introduction}
Pretrained contextualized text representation models have enabled massive advances in Natural Language Understanding (NLU) tasks, and achieved state-of-the-art performances in multiple NLP tasks \cite{howard2018universal,devlin2018bert}.  Early pretrained text representation models aimed at representing words by capturing their distributed syntactic and semantic properties using techniques like Word2vec~\cite{mikolov2013distributed} and GloVe~\cite{pennington2014glove}.  However, these models did not incorporate the context in which a word appears into its embedding.  This issue was addressed by generating contextualized representations using models like ELMO~\cite{peters2018deep}).

Recently, there has been a focus on applying transfer learning by fine-tuning large pretrained language models for downstream NLP/NLU tasks with a relatively small number of examples, resulting in notable performance improvement for these tasks.
This approach takes advantage of the language models that had been pre-trained in an unsupervised manner (or sometimes called self-supervised). However, this advantage comes with drawbacks, particularly the huge corpora needed for pre-training, in addition to the high computational cost of days needed for training (latest models required 500+ TPUs or GPUs running for weeks~\cite{conneau2019unsupervised,raffel2019exploring,adiwardana2020humanlike}).  These drawbacks restricted the availability of such models to English mainly and a handful of other languages.  To remedy this gap, multilingual models have been trained to learn representations for +100 languages simultaneously, but still fall behind single-language models due to little data representation and small language-specific vocabulary.  While languages with similar structure and vocabulary can benefit from the shared representations~\cite{conneau2019unsupervised}, this is not the case for other languages, like Arabic, which differ in morphological and syntactic structure and share very little with other abundant Latin-based languages. 

In this paper, we describe the process of pretraining the BERT transformer model~\cite{devlin2018bert} for the Arabic language, and which we name \textsc{AraBERT}.  We evaluate \textsc{AraBERT} on three Arabic NLU downstream tasks that are different in nature: \Ni Sentiment Analysis (SA), \Nii Named Entity Recognition (NER), and \Niii Question Answering (QA).  The experiments results show that \textsc{AraBERT} achieves state-of-the-art performances on most datasets, compared to several baselines including previous multilingual and single-language approaches.  The datasets that we considered for the downstream tasks contained both Modern Standard Arabic (MSA) and Dialectal Arabic (DA).

Our contributions can be summarized as follows:
\begin{itemize}
    \item A methodology to pretrain the BERT model on a large-scale Arabic corpus.
    \item Application of \textsc{AraBERT} to three NLU downstream tasks: Sentiment Analysis, Named Entity Recognition and Question Answering.
    \item Publicly releasing \textsc{AraBert} on popular NLP libraries.
\end{itemize}

The rest of the paper is structured as follows.  Section~\ref{sec:related_work} provides a concise literature review of previous work on language representation for English and Arabic.  Section~\ref{sec:methodology} describes the methodology that was used to develop \textsc{AraBERT}.  Section~\ref{sec:evaluation} describes the downstream tasks and benchmark datasets that are used for evaluation.  Section~\ref{sec:experiments} presents the experimental setup and discusses the results.  Finally, section~\ref{sec:conclusion} concludes and points to possible directions for future work.

\section{Related Works}
\label{sec:related_work}

\subsection{Evolution of Word Embeddings}
The first meaningful representations for words started with the word2vec model developed by \cite{mikolov2013distributed}. Since then, research started moving towards variations of word2vec like of GloVe \cite{pennington2014glove} and fastText \cite{mikolov2017advances}. While major advances were achieved with these early models, they still lacked contextualized information, which was tackled by ELMO \cite{peters2018deep}. The performance over different tasks improved noticeably, leading to larger structures that had superior word and sentence representations. Ever since, more language understanding models have been developed such as ULMFit \cite{howard2018universal}, BERT \cite{devlin2018bert}, RoBERTa \cite{liu2019roberta}, XLNet \cite{yang2019xlnet}, ALBERT \cite{lan2019albert}, and T5 \cite{raffel2019exploring}, which offered improved performance by exploring different pretraining methods, modified model architectures and larger training corpora.

\subsection{Non-contextual Representations for Arabic}
Following the success of the English word2vec \cite{mikolov2013distributed}, the same feat was sought by NLP researchers to create language specific embeddings. Arabic word2vec was first attempted by ~\cite{soliman2017aravec}, and then followed by a Fasttext model \cite{bojanowski2017enriching} trained on Wikipedia data and showing better performance than word2vec. To tackle dialectal variations in Arabic~\cite{erdmann2018addressing} presented techniques for training multidialectal word embeddings on relatively small and noisy corpora, while~\cite{abu-farha-magdy-2019-mazajak,abdul2018you} provided Arabic word embeddings trained on $\sim$250M tweets.

\subsection{Contextualized Representations for Arabic}
For non-English languages, Google released a multilingual BERT \cite{devlin2018bert} supporting 100+ languages with solid performance for most languages.
However, pre-training monolingual BERT for non-English languages proved to provide better performance than the multilingual BERT such as Italian BERT Alberto \cite{polignano2019alberto} and other publicly available BERTs~\cite{martin2019camembert,de2019bertje}. Arabic specific contextualized representations models, such as hULMonA \cite{eljundi2019hulmona}, used the ULMfit structure, which had a lower performance that BERT on English NLP Tasks.
\section{\textsc{AraBERT}: Methodology}
\label{sec:methodology}

In this paper, we develop an Arabic language representation model to improve the state-of-the-art in several Arabic NLU tasks. We create \textsc{AraBERT} based on the BERT model, a stacked Bidirectional Transformer Encoder~\cite{devlin2018bert}.  This model is widely considered as the basis for most state-of-the-art results in different NLP tasks in several languages. We use the BERT-base configuration that has 12 encoder blocks, 768 hidden dimensions, 12 attention heads, 512 maximum sequence length, and a total of $\sim$110M parameters\footnote{Further details about the transformer architecture can be found in~\cite{vaswani2017attention}}. We also introduced additional preprocessing prior to the model's pre-training, in order to better fit the Arabic language.  Below, we describe the pre-training setup, the pre-training dataset for \textsc{AraBERT}, the proposed Arabic-specific preprocessing, and the fine-tuning process.

\subsection{Pre-training Setup}
Following the original BERT pre-training objective, we employ the \emph{Masked Language Modeling} (MLM) task by adding whole-word masking where; 15\% of the $N$ input tokens are selected for replacement.  Those tokens are replaced 80\% of the times with the [MASK] token, 10\% with a random token, and 10\% with the original token.  Whole-word masking improves the pre-training task by forcing the model to predict the whole word instead of getting hints from parts of the word.  We also employ the \emph{Next Sentence Prediction} (NSP) task that helps the model understand the relationship between two sentences, which can be useful for many language understanding tasks such as Question Answering.

\subsection{Pre-training Dataset}
The original BERT was trained on 3.3B words extracted from English Wikipedia and the Book Corpus~\cite{zhu2015aligning}.  Since the Arabic Wikipedia Dumps are small compared to the English ones, we manually scraped Arabic news websites for articles.  In addition, we used two publicly available large Arabic corpora:
(1) the 1.5 billion words Arabic Corpus~\cite{el20161}, which is a contemporary corpus that includes more than 5 million articles extracted from ten major news sources covering 8 countries, and (2) OSIAN: the Open Source International Arabic News Corpus~\cite{zeroual-etal-2019-osian} that consists of 3.5 million articles ($\sim$1B tokens) from 31 news sources in 24 Arab countries.

The final size of the pre-training dataset, after removing duplicate sentences, is 70 million sentences, corresponding to $\sim$24GB of text.  This dataset covers news from different media in different Arab regions, and therefore can be representative of a wide range of topics discussed in the Arab world.  It is worth mentioning that we preserved words that include Latin characters, since it is common to mention named entities, scientific or technical terms in their original language, to avoid information loss.

\subsection{Sub-Word Units Segmentation}
\setcode{utf8}
Arabic language is known for its lexical sparsity which is due to the complex concatenative system of Arabic~\cite{al2017aroma}. Words can have different forms and share the same meaning.  For instance, while the definite article ``\RL{Al} - \textit{Al}'', which is equivalent to ``the'' in English, is always prefixed to other words, it is not an intrinsic part of that word.  Hence, when using a BERT-compatible tokenization, tokens will appear twice, once with ``\textit{Al-}'' and once without it.  For instance, both ``\RL{kitAb}- \emph{kitAb}'' and ``\RL{al-kitAb}-\emph{AlkitAb}'' need to be included in the vocabulary, leading to a significant amount of unnecessary redundancy.

To avoid this issue, we first segment the words using Farasa~\cite{abdelali2016farasa} into stems, prefixes and suffixes.  For instance, ``\RL{al-lo.gaT} - \textit{Alloga}'' becomes \RL{\RL{Al}\nospace+ \RL{lo.ga +T}} - \textit{Al+ log +a}''.  Then, we trained a SentencePiece (an unsupervised text tokenizer and detokenizer~\cite{kudo2018subword}), in unigram mode, on the segmented pre-training dataset to produce a subword vocabulary of $\sim$60K tokens.  To evaluate the impact of the proposed tokenization, we also trained SentencePiece on non-segmented text to create a second version of \textsc{AraBERT} (AraBERTv0.1) that does not require any segmentation.
The final size of vocabulary was 64k tokens, which included nearly 4K unused tokens to allow further pre-training, if needed.

\subsection{Fine-tuning}
\paragraph{Sequence Classification}
To fine-tune AraBERT for sequence classification, we take the final hidden state of the first token, which corresponds to the word embedding of the special ``[CLS]" token prepended to the start of each sentence. We then add a simple feed-forward layer with standard Softmax to get the probability distribution over the predicted output classes. During fine-tuning, the classifier and the pre-trained model weights are trained jointly to maximize the log-probability of the correct class.
\paragraph{Named Entity Recognition}
For the NER task, each token in the sentence is labeled with the IOB2 format~\cite{ratnaparkhi1998maximum}, where the ``B'' tag corresponds to the first word of the entity, the ``I'' tag corresponds to the rest of the words of the same entity, and the ``O'' tag indicates that the tagged word is not a desired named entity. Hence, we treat the system as a multi-class classification process, which allows us to use some text classification methods to label the tokens. Furthermore, after using the AraBERT tokenizer, we only input the first sub-token of each word to the model.

\paragraph{Question Answering}
In the QA, given a question and a passage containing the answer, the model needs to select a span of text that contains the answers. This is done by predicting a ``start" token and an ``end" token on condition that the ``end" token should appear after the ``start" token. During training, the final embedding of every token in the passage is fed into two classifiers, each with a single set of weights, which are applied to every token. The dot product of the output embeddings and the classifier is then fed into a softmax layer to produce a probability distribution over all the tokens. The token with the highest probability of being a ``start" toke is then selected, and the same process is repeated for the ``end" token.
\section{Evaluation}
\label{sec:evaluation}

We evaluated \textsc{AraBERT} on three Arabic language understanding downstream tasks: Sentiment Analysis, Named Entity Recognition, and Question Answering.  As a baseline, we compared \textsc{AraBERT} to the multilingual version of BERT, and to other state-of-art results on each task.

\subsection{Sentiment Analysis}
We evaluated \textsc{AraBERT} on the following Arabic sentiment datasets that cover different genres, domains and dialects.

\begin{itemize}
\item \textbf{HARD:} The Hotel Arabic Reviews Dataset~\cite{elnagar2018hotel} contains 93,700 hotel reviews written in both Modern Standard Arabic (MSA) and in dialectal Arabic.  Reviews are split into positive and negative reviews, where a negative review has a rating of 1 or 2, a positive review has a rating of 4 or 5, and neutral reviews with rating of 3 were ignored. 
\item \textbf{ASTD:}
The Arabic Sentiment Twitter Dataset~\cite{nabil-etal-2015-astd} contains 10,000 tweets written in both MSA and Egyptian dialect.  We tested on the balanced version of the dataset, referred to as ASTD-B.
\item \textbf{ArSenTD-Lev:}
The Arabic Sentiment Twitter Dataset for LEVantine~\cite{baly2018arsentd} contains 4,000 tweets written in Levantine dialect with annotations for sentiment, topic and sentiment target.  This is a challenging dataset as the collected tweets are from multiple domains and discuss different topics.
\item \textbf{LABR:}
The Large-scale Arabic Book Reviews dataset~\cite{aly-atiya-2013-labr} contains 63,000 book reviews written in Arabic.  The reviews are rated between 1 and 5.  We benchmarked our model on the unbalanced two-class dataset, where reviews with ratings of 1 or 2 are considered negative, while those with ratings of 4 or 5 are considered positive.
\item \textbf{AJGT:}
The Arabic Jordanian General Tweets dataset~\cite{alomari2017arabic} contains 1,800 tweets written in Jordanian dialect. The tweets were manually annotated as either positive or negative.
\end{itemize}

\paragraph{Baselines:} Sentiment Analysis is a popular Arabic NLP task.  Previous approaches relied on sentiment lexicons such as ArSenL~\cite{badaro2014large}, which is a large-scale lexicon of MSA words that is developed using the Arabic WordNet in combination with the English SentiWordNet.  Recurrent and recursive neural networks were explored with different choices of Arabic-specific processing~\cite{al2015deep,al2017aroma,baly2017sentiment}.  Convolutional Neural Networks (CNN) were trained with pre-trained word embeddings~\cite{dahou2019arabic}.  A hybrid model was proposed by~\cite{abu-farha-magdy-2019-mazajak}, where CNNs were used for feature extraction, and LSTMs were used for sequence and context understanding.  Current state-of-the-art results are achieved by the hULMonA model~\cite{eljundi2019hulmona}, which is an Arabic language model that is based on the ULMfit architecture~\cite{howard2018universal}.  We compare the results of \textsc{AraBERT} to those of hULMonA.

\subsection{Named Entity Recognition}
This task aims to extract and detect named entities in the text.  It is framed as a word-level classification (or tagging) task, where the classes correspond to pre-defined categories such as names, locations, organizations, events and time expressions.  For evaluation, we use the Arabic NER corpus (ANERcorp)~\cite{benajiba2007anersys}.
This dataset contains 16.5K entity mentions distributed among 4 entities categories, \emph{person} (39\%), \emph{organization}: (30.4\%), \emph{location}: (20.6\%), and \emph{miscellaneous}: (10\%).

\paragraph{Baselines:} Advances in the NER task have been focusing on English, namely on the CoNLL 2003~\cite{sang2003introduction} dataset. Initially, NER was tackled with Conditional Random Fields (CRF)~\cite{lafferty2001conditional}. Later on, CRFs were used on top of Bi-LSTM models~\cite{huang2015bidirectional,lample2016neural} presenting significant improvements over standalone CRFs. Bi-LSTM-CRF structures were then used with contextualized  embeddings that displayed further improvements~\cite{peters2018deep}.
Lastly, large pre-trained transformers showed slight improvement, setting the current state-of-the-art performance~\cite{devlin2018bert}.  As for Arabic, We compare \textsc{AraBERT} performance with Bi-LSTM-CRF baseline that set the previous state-of-the-art performance~\cite{el2019arabic}, and with BERT multilingual.

\subsection{Question Answering}
Open-domain Question Answering (QA) is one of the goals of artificial intelligence, this goal can be achieved by leveraging natural language understanding and knowledge gathering~\cite{47761}. English QA research has been fueled by the release of large datasets such as Stanford Question Answering Dataset (SQuAD)~\cite{rajpurkar2016squad}. On the other hand, research in Arabic QA has been hindered by the lack of such massive datasets, and by the fact that Arabic presents its own challenges such as:
\begin{itemize}
\setlength\itemsep{-1em}
    \item Inconsistent name spelling (ex: Syria in Arabic can be written as ``\RL{sOriyA} - \textit{sOriyA}" and ``\RL{sOriyT} - \textit{sOriyT}" )
    \item Name de-spacing (ex: The name is written as ``\RL{`bdAl`azIz} - \textit{AbdulAzIz}" in the question, and  ``\RL{`bd Al`azIz} - \textit{Abdul AzIz}" in the answer)
    \item Dual form ``\RL{Almo_tan_A}", which can have multiple forms (ex: ``\RL{qalamAn}" - ``\textit{qalamAn}" or ``\RL{qalamyn}" - ``\textit{qalamyn}" meaning ``\textit{two pencils}")
    \item Grammatical gender variation: all nouns, animate and inanimate objects are classified under two genders either masculine or feminine (ex: ``\RL{kabIr}" - ``\textit{kabIr}" and ``\RL{kabIrT}" - ``\textit{kabIrT}"
\end{itemize}

We evaluate \textsc{AraBERT} on the Arabic Reading Comprehension Dataset (ARCD)~\cite{mozannar2019neural}
, where the task is to find the span of the answer in a document for a given question.
ARCD contains 1395 questions on Wikipedia articles along with 2966  machine translated questions and answers from the SQuAD dubbed (Arabic-SQuAD). We train on the whole Arabic-SQuAD and on 50\% of ARCD and test on the remaining 50\% of ARCD.

\paragraph{Baselines} Multilingual BERT had previously achieved state of the art results on ARCD.

\section{Experiments}
\label{sec:experiments}

\subsection{Experimental Setup}
\paragraph{Pretraining}
In our experiments, the original implementation of BERT on TensorFlow was used.  The data for pre-training was sharded, transformed into TFRecords, and then stored on Google Cloud Storage.  Duplication factor was set to 10, a random seed of 34, and a masking probability of 15\%. The model was pre-trained on a TPUv2-8 pod for 1,250,000 steps.  To speed up the training time, the first 900K steps were trained on sequences of 128 tokens, and the remaining steps were trained on sequences of 512 tokens. The decision of stopping the pre-training was based on the performance of downstream tasks. We follow the same approach taken by the open-sourced German BERT~\cite{deepset}. Adam optimizer was used, with a learning rate of 1e-4, batch size of 512 and 128 for sequence length of 128 and 512 respectively. Training took 4 days, for 27 epochs over all the tokens.

\paragraph{Fine-tuning}
Fine-tuning was done independently using the same configuration for all tasks. We do not run extensive grid search for the best hyper-parameters due to computational and time constraints. We use the splits provided by the dataset's authors when available. and the standard 80\% and 20\% when not\footnote{The scripts used to create the datasets are available on our Github repo \url{https://github.com/aub-mind/arabert}}.

\renewcommand{\tabularxcolumn}[1]{m{#1}}

\setcode{utf8}
\subsection{Results}

Table~\ref{tbl:results} illustrates the experimental results of applying AraBERT to multiple Arabic NLU downstream tasks, compared to state-of-the-art results and the multilingual BERT model (mBERT).

\paragraph{Sentiment Analysis}
For Arabic sentiment analysis, the results in Table~\ref{tbl:results} show that both versions of AraBERT outperform mBERT and other state-of-the-art approaches on most tested datasets.  Even though AraBERT was trained on MSA, the model was able to preform well on dialects that were never seen before. 

\begin{table}[!ht]
\begin{center}
\centering
\caption{Performance of AraBERT on Arabic downstream tasks compared to mBERT and previous state of the art systems\label{tbl:results}}

\resizebox{0.5\textwidth}{!}{
    \begin{tabular}{l | c | c c c}
    \toprule
    \textbf{Task}               & \textbf{metric} & \textbf{prev. SOTA} & \textbf{mBERT} & \textbf{AraBERTv0.1/ v1} \\
    \toprule
    \textbf{SA (HARD)}          & Acc. & 95.7*  & 95.7 & \textbf{96.2} / 96.1 \\
    \textbf{SA (ASTD)}          & Acc. & 86.5*  & 80.1 & 92.2 / \textbf{92.6} \\
    \textbf{SA (ArsenTD-Lev)}   & Acc. & 52.4*  & 51.0 & 58.9 / \textbf{59.4} \\
    \textbf{SA (AJGT)}          & Acc. & 92.6** & 83.6 & 93.1 / \textbf{93.8} \\
    \textbf{SA (LABR)}          & Acc. & \textbf{87.5}$^{\dagger}$ & 83.0 & 85.9 / 86.7 \\
    \midrule
    \textbf{NER (ANERcorp)}     & macro-F1 & 81.7${^{\dagger}}{^{\dagger}}$ & 78.4 & \textbf{84.2} / 81.9 \\
    \midrule
                                & Exact Match & & \textbf{34.2} & 30.1 / 30.6 \\
    \textbf{QA (ARCD)}          & macro-F1 & mBERT & 61.3 &61.2 / \textbf{62.7} \\
                                & Sent. Match & & 90.0 & \textbf{93.0} / 92.0 \\
    \bottomrule
    \multicolumn{5}{l}{*~\cite{eljundi2019hulmona}} \\
    \multicolumn{5}{l}{**~\cite{dahou2019multi}} \\
    \multicolumn{5}{l}{${^{\dagger}}$~\cite{dahou2019multi}} \\
    \multicolumn{5}{l}{${^{\dagger}}{^{\dagger}}$~Previous state of the art performance by BiLSTM-CRF model} \\
    \end{tabular}
}
\end{center}
\end{table}

\paragraph{Named Entity Recognition}
Results in Table~\ref{tbl:results} show that AraBERTv0.1 improved results by 2.53 points in F1 score scoring 84.2 compared with the Bi-LSTM-CRF model, making AraBERT the new state-of-the-art for NER on ANERcorp. Testing AraBERT with tokenized suffixes and prefixes showed results similar to that of the Bi-LSTM-CRF model. We believe that the reason this happened is that the start token (B-label) is referenced to the suffixes most of the time. An example of this, ``\RL{AljAmi`T}'' with a label B-ORG becomes ``\RL{Al}'', ``\RL{jAmi`T}'' with labels B-ORG, I-ORG respectively, providing misleading starting cues to the model. Testing multilingual BERT, it proved inefficient as we got results lower than the baseline model.

\paragraph{Question Answering}
While the results in Table~\ref{tbl:results} show an improvement in F1-score, the exact match scores were significantly lower.
Upon further examination of the results, the majority of the erroneous answers differed from the true answer by one or two words with no significant impact on the semantics of the answer.  Examples are shown in Tables~\ref{arcd_example:1} and~\ref{arcd_example:2}.  We also report a 2\% absolute increase in the sentence match score over mBERT, which is the previous state-of-the-art.  Sentence Match (SM) measures the percentage of predictions that are within the same sentence as the ground truth answer.

\begin{table}[!ht]
\centering
\caption{Example of an erroneous results from the ARCD test set: the only difference is the preposition ``\RL{fI} - \textit{In}".\label{arcd_example:1}}
\resizebox{0.5\textwidth}{!}{%
    \begin{tabular}{c|r}
        \toprule
        \textbf{Question} & \RL{"Ayn t'asast mona.zamaT Al-'omam Al-mota.hidT?} \\ & \textit{where was the united nations established?} \\
        \midrule
        \textbf{Ground Truth} & \textit{In San Francisco} -- \RL{fI sAn fransIskO} \\
        \textbf{Predicted Answer} & \textit{San Francisco} -- \RL{sAn fransIskO} \\
        \bottomrule
    \end{tabular}
}
\end{table}

\begin{table}[!ht]
\centering
\caption{Another example of an erroneous results from the ARCD test set: the predicted answer does not include ``introductory'' words.\label{arcd_example:2}}
\resizebox{0.5\textwidth}{!}{%
    \begin{tabular}{c|r}
        \toprule
        \textbf{Question} & \RL{mA hO Al-n.zAm Al-_hA.s bdOlT Al-namsA?} \\ & \textit{What is the type of government in Austria?} \\
        \midrule
        \textbf{Ground Truth} & \textit{Austria is a federal republic} -- \RL{Al-namsA hy jomhOrYT fIdirAliyT}  \\
        \textbf{Predicted Answer} & \textit{A federal republic} -- \RL{jomhOrYT fIdirAliyT} \\
        \bottomrule
    \end{tabular}
}
\end{table}

\subsection{Discussion}
AraBERT achieved state-of-the-art performance on sentiment analysis, named entity recognition, and the question answering tasks. This adds truth to the assumption that pre-trained language models on a single language only surpass the performance of a multilingual model. This jump in performance has many explanations. First, data size is a clear factor for the boost in performance. AraBERT used around 24GB of data in comparison with the 4.3G Wikipedia used for the multilingual BERT. Second, the vocab size used in the multilingual BERT is 2k tokens in comparison with 64k vocab size used for developing AraBERT. Third, with the large data size, the pre-training distribution has more diversity. As for the fourth point, the pre-segmentation applied before BERT tokenization improved performance on SA and QA tasks but reduced it on the NER task.
It is also noted that the pre-processing applied to the pre-training data took into consideration the complexities of the Arabic language. Hence, increased the effective vocabulary by excluding unnecessary redundant tokens that come with certain common prefixes, and help the model learn better by reducing the language complexity.
We believe these factors helped to reach state-of-the-art results on 3 different tasks and 8 different datasets. Obtained results indicate that the advantage we got in the datasets considered are better understood in a monolingual model than of a general language model trained on Wikipedia crawls such as multilingual BERT.

\section{Conclusion}
\label{sec:conclusion}
AraBERT sets a new state-of-the-art for several downstream tasks for Arabic language. It is also  300MB smaller than multilingual BERT. By publicly releasing our AraBERT models, we hope that it will be used to serve as the new baseline for the various Arabic NLP tasks, and hope that this work will act as a footing stone to building and improving future Arabic language understanding models. We are currently working on publishing an AraBERT version that won't depend on external tokenizers. We are also in the process of training models with a better understanding of the various dialects that the Arabic language has across different Arabic countries.
\section{Acknowledgments}
We would like to express special thanks to Dr.~Ramy Baly (Massachusetts Institute of Technology)  for the useful discussions and suggestions, to Dr.~Dirk Goldhahn (Universität Leipzig) for access to the OSIAN dataset, to TFRC for the free access to cloud TPUs, and to As-Safir newspaper, and Yakshof for providing us with their news articles.
\section{References}
\bibliographystyle{lrec}
\bibliography{references}

\label{lr:ref}
\bibliographystylelanguageresource{lrec}

\end{document}